\pdfoutput=1
\documentclass{article}
\usepackage[preprint]{nips_2018}

\usepackage[utf8]{inputenc} %
\usepackage[T1]{fontenc}    %
\usepackage{hyperref}       %
\usepackage{url}            %
\usepackage{booktabs}       %
\usepackage{amsfonts}       %
\usepackage{nicefrac}       %
\usepackage{microtype}      %

\usepackage{bm}%
\usepackage{amsmath}
\usepackage{algorithm} 
\usepackage{algpseudocode}

\usepackage{graphicx}
\usepackage{subfigure}

\usepackage{url}
\usepackage{color}
\usepackage{mathrsfs}
\usepackage{amsthm}
\theoremstyle{definition}
\newtheorem{theorem}{Theorem}
\newtheorem*{theorem*}{Theorem}
\newtheorem{definition}{Definition}
\newtheorem*{definition*}{Definition}

\newtheorem*{lemma*}{Lemma}

\newtheorem*{prop*}{Proposition}
\newtheorem{corollary}[theorem]{Corollary}
\DeclareMathOperator*{\argmin}{arg\,min}
\DeclareMathOperator*{\argmax}{arg\,max}

\title{Reconnaissance and Planning algorithm for constrained MDP}

\author{
  Shin-ichi~Maeda\\
  Preferred Networks, Inc.\\
  \texttt{ichi@preferred.jp} \\
 \And
   Hayato~Watahiki
     \thanks{Work done while Hayato Watahiki worked at Preferred Networks, Inc., Tokyo, Japan.}\\
  The University of Tokyo\\
  \texttt{watahiki@eidos.ic.i.u-tokyo.ac.jp} 
   \And
   Shintarou~Okada \\
   Preferred Networks, Inc.\\
  \texttt{okada@preferred.jp} 
   \And
   Masanori~Koyama \\
   Preferred Networks, Inc.\\
  \texttt{masomatics@preferred.jp} 
}

\begin{document}

\maketitle

\begin{abstract}
Practical reinforcement learning problems are often formulated as constrained Markov decision process (CMDP) problems, in which the agent has to maximize the expected return while satisfying a set of prescribed safety constraints.
In this study, we propose a novel simulator-based method to approximately solve a CMDP  problem without making any compromise on the safety constraints. 
We achieve this by decomposing the CMDP into a pair of MDPs; reconnaissance MDP and planning MDP.  
The purpose of \textit{reconnaissance} MDP is to evaluate the set of actions that are safe, and the purpose of \textit{planning} MDP is to maximize the return while using the actions authorized by \textit{reconnaissance} MDP. 
RMDP can define a set of safe policies for any given set of safety constraint, and this set of safe policies can be used to solve another CMDP problem with different reward.
Our method is not only computationally less demanding than the previous simulator-based approaches to CMDP, but also capable of finding a competitive reward-seeking policy in a high dimensional environment, including those involving multiple moving obstacles. 
\end{abstract}

\section{Introduction}

With recent advances in reinforcement learning (RL), it is becoming possible to learn complex reward-maximizing policy in an increasingly more complex environment
\cite{Mnih15Atari, Silver16Go, Andrychowicz18dexterous_Hand, James18Sim2Real, Kalashnikov18QT-opt}.
However, not all policies found by standard RL methods are physically \textit{safe} in real-world applications, and a naive application of RL can lead to catastrophic results.  
This has long been one of the greatest challenges in the application of reinforcement learning to mission-critical systems. 
In a popular setup, one assumes a Markovian system together with a predefined set of  dangerous states that must be avoided, and formulates the problem as a type of constrained Markov decision process (CMDP) problem.  That is,
based on the classical RL notations in which $\pi$ represents a policy of the agent, we aim to solve 
\begin{eqnarray}
\max _{\pi} E^{\pi}[R(h)] ~~~ {\rm{s.t.}}~~~E^{\pi}[D(h)] \leq c,
\label{eq:CMDP}
\end{eqnarray}
where $h$ is a trajectory of state-action pairs, $R(h)$ is the total return that can be obtained by $h$, and $D(h)$ is the measure of how dangerous the trajectory $h$ is.
Most methods of reinforcement learning solves the optimization problem about $\pi$ by a sequence of iterative updates.  
The difficulty of CMDP problem lies in the evaluation of the safeness of the $\pi$ suggested at every update.
The evaluation of the safeness requires the evaluation of integrals with respect to future possibilities, whose cardinality increases exponentially with the length of the future and the number of randomly moving objects in the environment.

Lagrange multiplier-based methods \cite{Altman99constrainedMDP, Geibel05RiskSensitiveRL} tackle this problem by aiming to satisfy the constraint softly, and provide the guarantee that the obtained solution is safe if \textit{optimal} lambda is chosen. 
Trust region optimization (TRO) \cite{Achiam17ConstrainedPO, Chow19Lyapunov}, and the methods based on Lyapunov function \cite{Chow18Lyapunov, Chow19Lyapunov} take the approach of constructing at each update step a pool of policies that are most likely safe.
The precise construction of safe pool and the finding of optimal hyper-parameter, however, are computationally heavy tasks in high-dimensional state spaces, and a strong regularity assumption about the system becomes necessary in using these methods in practice.

Presence of a good simulator is particularly important for safe applications when the event to be avoided is a ``rare" catastrophic accident,  because an immense number of samples will be required to collect information about the cause of the accident. 

Model Predictive Control (MPC) is perhaps the oldest family of simulator-based methods \cite{Falcone07MPC, Wang10FastMPC, Cairano13MPC, Weiskircher17MPC} for carrying out tasks under safe constraints. 
Model Predictive Control uses the philosophy of receding horizon and predicts the future outcome of actions in order to determine what action the agent should take in the next step.
If the future-horizon to consider is sufficiently short and the dynamics is deterministic, the prediction can often be approximated well by linear dynamics, which can be evaluated instantly. 
Because MPC must finish its assessment of the future before taking an action, its performance is limited by the speed of the predictions.
If only a short horizon is taken into account, MPC may suggest a move to a state leading to a catastrophe.

\begin{figure}
\includegraphics[width=140mm]{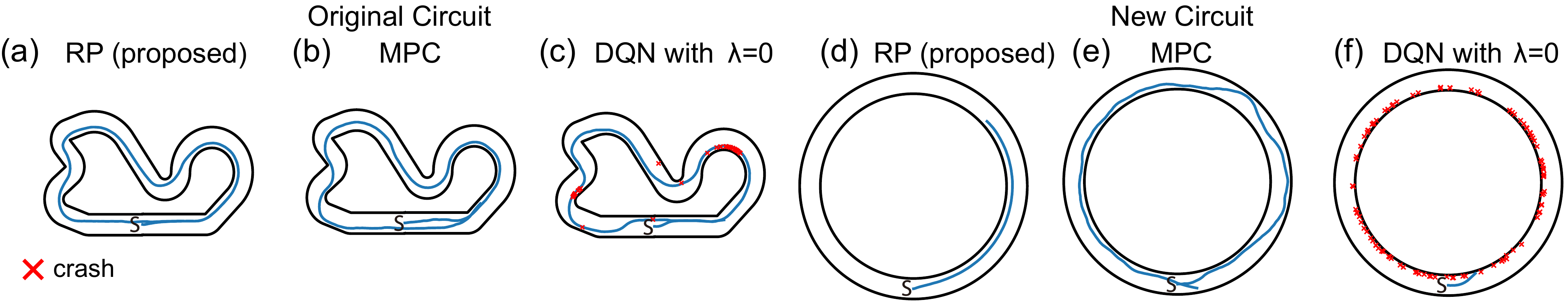}
\caption{The trajectories produced by the the policy trained by our proposed method ((a) and (d)), 4-step MPC ((b), (e)), the policy trained with penalized DQN ((c) and (f)). The trajectories on circular circuit were produced by the policies trained on the original circuit. $S$ represents the initial position of the agent. 
The red marks represents the places at which the agent crashed into the wall.  Our method can maneuver through the environment without any accident.
The policy of DQN cannot adapt to new environment because it is not aware of the new location of the wall in the new circuit.
Meanwhile, the policy found by our method can skirt the danger in new environment because it is finding a policy from a set of policies defined by a reward independent threat function constructed just for the sake of safety.
MPC can also finish a lap without any accidents. MPC, however, takes an order of magnitude more computation time than our method (See Fig. \ref{Circuit2}).
} \label{fig:route}
\end{figure}

In this study, we propose a novel simulator-based approach that looks for a solution of a CMDP problem by decomposing the CMDP into a pair of MDPs: a reconnaissance MDP (R-MDP) and planning MDP (P-MDP).
The purpose of R-MDP is to (1) \textit{recon} the state space, (2) evaluate the \textit{threat} function that measures the potential danger at each state, and (3) construct a pool of policies that are safe in the sense of satisfying a user-specified constraint. 
After solving the R-MDP problem, we solve the P-MDP problem consisting of the original MDP while restricting our policy-search to the R-MDP specified pool of safe policies. 
If we can find one safe policy, we can use the threat function construct non-empty set of policies that are guaranteed to be safe. %
The threat function we compute in R-MDP is mathematically close to the Lyapunov function considered previously by Chow et al. \cite{Chow18Lyapunov,Chow19Lyapunov}. 
However, unlike these prior works, we do not have to evaluate the safety of a given policy more than once.
Because our method is computationally light, it can be used to solve CMDP problems in high-dimensional spaces with relative ease. 
Fig. \ref{fig:route} illustrates the routes on the circuit taken by agents trained with various methods of CMDP and the locations of accidents made by the agents. 
The agent trained with our algorithm is finding a safe and efficient route.

The advantages of our approach are multifold. 
Because the threat function alone specifies the pool of safe policies in our framework, we can re-use the pool specified by the R-MDP constructed for one CMDP problem to solve another CMDP problem with a different reward function and the same safety constraint. 
Our formulation of the threat function can also be used to solve a MDP problem with a constraint on the probability of catastrophic failure.
By applying a basic rule of probability measure to a set of threat functions, we can solve a CMDP problem with multiple safety constraints as well.
This allows us to find a good reward-seeking policy for a sophisticated task like \textit{safely navigating through a crowd of randomly moving objects}.
Although our method does not guarantee to find the optimal solution of the CMDP problem, our method prioritizes safety and is still able to find a safe policy that is competitive in terms of reward-seeking ability. 
To the best of our knowledge, there has not been any study to date that has succeeded in solving a CMDP in dynamical environments as high-dimensional as the ones discussed in this study. 

The work that is algorithmically closest to our approach is \cite{Bouton18safe}, which computes for each state the state-dependent set of actions $\{a(s);  \max_{\pi} P^{\pi}(\textrm{constraint is satisfied})| s,a ) > \lambda \} $ that guarantees the safety when the constraint can be written as Linear Temporal Logic. 
This paper provides similar approach for the form of risk that is more commonly used in the field of safe-reinforcement learning. 
The paper also provides more solid theoretical justification to the safety-guarantee. 
The following list summarizes the advantages of our new framework. %
\begin{itemize}
  \setlength{\parskip}{0cm} 
  \setlength{\itemsep}{1mm} 
    \item  The R-MDP problem of identifying the set of safe policies needs to be solved only once.
    Moreover, one does not necessarily need to obtain the absolute optimal solution for the R-MDP problem in order to find a good reward seeking safe-policy from the ensuing C-MDP problem.%
    \item The policy proposed by our method is almost always safe.  
    If we can find a safe policy from the RMDP problem, we can always guarantee the safety.
    \item The threat function evaluated by the R-MDP can be re-used for another CMDP problem with safety constraints on the same quantities.  %
    \item The P-MDP can be solved with or without access to a simulator.
\end{itemize}

\section{Method}
%

\subsection{Problem formulation and setup}
We begin this section with the notations and assumptions that we are going to use throughout the paper. 
We assume that the system in consideration is a discrete-time constrained Markov Decision Process with finite horizon, defined by a tuple
$(S, A, r, d, P, P_0)$, where $S$ is the set of states, $A$ is the set of actions,
$P(s,a,s')$ is the density of the state transition probability from $s$ to $s'$ when the action is $a$,  $r(s,a)$ is the reward obtained by action $a$ at state $s$,  $d(s,a)$ is the non-negative danger of taking action $a$ at state $s$, and $P_0$ is the distribution of the initial state.
We use $\pi(a|s)$ to denote the policy $\pi$'s probability of taking an action $a$ at a state $s$.
Also, for ease of notation, we use 
$r$ to denote $r(s,a)$, and $r_{t+1}$ to denote $r(s_t, a_t)$. 
Likewise, we will use $d$ and $d_{t+1}$ to  denote $d(s,a)$ and $d(s_t, a_t)$ respectively. 
Finally, for an arbitrary set $U$, we  will use $U^c$ to denote its complement.  

Next, we present the optimization problem \eqref{eq:CMDP}  in more formality.
The ultimate goal of CMDP(Constrained Markoc Decision Process Problem) is to find the policy $\pi^*$ that solves 
\begin{align}
 \argmax_{\pi} E_{\pi}\left[\textstyle \sum_{k=1}^{T} \gamma^k r_k    \right], \; %
 {\rm{s.t. }} E_{\pi}\left[\textstyle \sum_{k=1}^{T} \beta^k d_k \right] \leq c,
\label{eq: formal CMDP}
\end{align}
where $c\geq 0$, $\gamma, \beta \in [0,1)$ and $E_{\pi}[\cdot]$ denotes the expectation
with respect $\pi$, $P$ and $P_0$. 
Unless otherwise denoted, we will use $E_\pi$ to refer to the integration with respect to both $\pi$ and $P_0$.    
In our formulation, 
we use the following \textit{threat function} as a danger-analogue of the action-value function. 
We define the threat function for a policy $\eta$ at $(s,a,t)$ by 
\begin{align}
    \mathscr{T}_t^\eta(s, a) = E_\eta \left[ \sum_{k=t+1}^{T} \beta^k d_k \mid  s_t = s, a_t = a\right]. \label{eq:def risk function}
\end{align} 
Informally, we can think of $\mathscr{T}_t^{\pi}(s,a)$ as the aggregated measure of threat that the agent with policy $\pi$ must face after taking the action $a$ in the state $s$ at time $t$. 
We may say that a policy $\pi$ is safe if $E_\pi \left[ \mathscr{T}_0^{\pi}(s,a)\right] \leq c$. 
To reiterate, out strategy is to (1) evaluate the threat function for a baseline policy, (2)  construct a pool of safe policy using the threat function, and (3) to look for a reward-seeking policy in the pool of the safe policies.
Before we proceed further, we describe several key definitions and lemmas that stem from the definition of threat function.

\subsection{Properties of threat functions and $\eta$ secure policies} \label{sec:method_core}
For now, let us consider a time-dependent safety threshold $x_t$ defined at each time $t$, and let $\eta$ be a baseline policy.
Then the set of $(\eta,\mathbf{x})$-secure actions the set of actions that are deemed \textit{safe} by $\eta$ for for risk threshold $x$ in the sense that agent's safety is guaranteed if it follows $\eta$ afterward.
\begin{definition}[$(\eta,\bm x)$-secure actions]
\begin{align}
A^{\eta, \mathbf{x}}(s) = \bigcap_{t \in \{0,\cdots,T\}} A^{\eta, \mathbf{x}}(s,t),
\end{align}  \label{def:secureActions}  %
where $A^{\eta, \mathbf{x}}(s,t) = \Bigl\{a ;  \mathscr{T}_t^\eta(s, a)\leq x_t\Bigr\}$, and $x_t$ is a non-negative time-dependent constant.
\end{definition}
Let $\prod _A = \{\pi;\mathrm{supp}(\pi) \subseteq A \}$. 
Then $\prod_{A^{\eta, x}}$ is a set of actions that very much represents the agent's freedom in seeking reward when following the policy $\eta$. 
But indeed, this set of actions is not always non-empty.
Let us define $(\eta, x)$-secure states $S^{\eta,\mathbf{x}}= \Bigl\{s\in S ; 
A^{\eta, \mathbf{x}}(s) \neq \emptyset \Bigr\}$ to be the set of states for which there is non-empty $(\eta,\bm x)$-secure actions.
Over such set of states, we want the agent to take actions that are safer than $\eta$. 
Let 
$$\mathcal{F}^{\eta}(s) = \{p(\cdot | s);  \forall t \in \{0,\cdots,T\}, E_{p}\left[\mathscr{T}^{\eta}_t(s, a)\right]
\leq E_{\eta}\left[\mathscr{T}^{\eta}_t(s, a)\right] \}.$$
We are going to use the following set of policies as the first candidate of pool from which to look for a reward-seeking safe policy:
\begin{align}
 {\textstyle \prod^{\eta, \mathbf{x}}} := & \{\pi ;  \pi(\cdot | s) \in {\textstyle \prod _{A^{\eta, \mathbf{x}}(s)}} \ {\rm{ at }} \ s \in S^{\eta,\mathbf{x}}, \ 
 {\rm{ otherwise }} \ \pi(\cdot | s) \in \mathcal{F}^{\eta}(s)  \}.
\end{align}
We will refer this policy as the set of $(\eta,\mathbf{x})$-secure policies.
Intuitively, this set shall increase as $\eta$ becomes safer. 
While this intuition unfornately does not  hold in general, it holds for its lower bound subset: 
\begin{align}
{\textstyle \prod^{\eta, \mathbf{x}}_1(s) }:=  \{\pi ;  \pi(\cdot | s) \in {\textstyle \prod _{A^{\eta, \mathbf{x}}(s)}} \ {\rm{ at }} \ s \in S^{\eta,\mathbf{x}}, \  {\rm{ otherwise }} \ \pi(\cdot | s) =  1_{\argmin_a E_{\eta}\left[\mathscr{T}^{\eta}_t(s, a)\right]}  \}
\end{align}
That is, $\prod_1^{\eta, x}(s) \subseteq \prod _1^{\eta', x}(s)$ whenever $\mathscr{T}_t^{\eta'}(s, a)\leq 
\mathscr{T}_t^\eta(s, a)$ for all $a$.

We are still not yet done. 
Up until here, we have been defining the set of actions based on $\eta$-defined measure of safety. 
As we will be using a policy other than $\eta$ to maximize the reward, we must take into account the risk that will be incurred in taking an action from a policy other than the one used for determining its risk:
\begin{theorem} \label{lemma:threat bound}
Let $d_{TV}(p,q)$ be the total variation distance\footnote{Total variation distance is defined as $d_{TV}(p(a),q(a)) = \frac{1}{2}\sum_a |p(a)-q(a)|$.} between two distributions $p$ and $q$.   
For a given policy $\eta$, let $\pi$ be a policy such that $\pi \in \prod^{\eta, x}$.
If  $ \mathscr{T}_0^\eta(s_0, a)\leq 
x_0$, then 
\begin{align} 
\begin{split}
\mathscr{T}_0^\pi(s_0) \leq x_0 + \sum_{t=1}^{T-1} \beta^t x_t E_{\pi}\left[ z_t(s_{t-1},a_{t-1})\mid s_0 \right],
\end{split}\label{eq:safethm}
\end{align}
where $z_t(s_{t-1},a_{t-1}) = E\left[ 1_{s_t \in S^{\eta}} d_{TV}(\pi(\cdot|s_t)  , \eta(\cdot|s_t))  \mid  s_{t-1}, a_{t-1} \right]$. 
\end{theorem} \label{thm:safe} 
For the proof, see the appendix. 
The bound \eqref{eq:safethm} in its raw form is not too useful because the  RHS depends on $\pi$ and the threat of $\pi$ is bounded implicitly. 
However, if appeal to the trivial upperbound for the total variation distance and set $x_t=c$,  we can achieve $\mathscr{T}^{\pi}_{0}(s_{0}) \leq c \left(1 + \sum_{t=1}^{T-1}\beta^t z_t \right) \leq c \left(1 + \sum_{t=1}^{T-1}\beta^t \right)$. 
The summation term in parenthesis is the very penalty that the agent must pay in taking action other than $\eta$, the safety-evaluating policy.
We can thus guarantee $\mathscr{T}^{\pi}_{0}(s_{0}) \leq c$ by just setting $x_t$ to a value smaller than $c$:   
\begin{corollary}
Let $x^*_c  = c \left(1 + \sum_{t=1}^{T-1} \beta^t \right)^{-1}$. Then 
$\pi \in \prod^{\eta, x^*_c}_1$ is safe.   \label{ref:thm2}
\end{corollary}
Thus, from any baseline policy $\eta$ satisfying $\mathscr{T}^\eta(s_0) \leq  x_c^*$, we can construct a pool of absolutely safe policies whose membership condition is based \textit{explicitly} on $\eta$ and $c$ alone.    
That the threshold expression in Eq. \eqref{ref:thm2} is free of $\pi$ is what allows us to decompose the CMDP problem into two separate MDP problems.
We can seek a solution to the CMDP problem by (1) looking for an $\eta$ satisfying $\mathscr{T}^\eta(s_0) \leq  x_c^*$, and (2) looking for the reward maximizing policy in $\pi \in \prod^{\eta, x^*_c}_1$.
We address the first problem by R-MDP, and the second problem by P-MDP.

Now, several remarks are in order.
First, if we take the limit of $\beta \to 0$, the $x^*_c$ in the above statement will approach $c$, and this just gives us the requirement that $\eta$ itself must be safe in order for $\prod^{\eta}$ to serve as a pool of safe policies.   %
Next, if we set $\beta =1$, then  $x^*_c \to 0$ as $T \to \infty$. 
This is in agreement with the law of large numbers; that is, any accident with positive probability is bound to happen at some point.
Also, recall that we have 
$\prod_1^{\eta, x}(s) \subseteq \prod _1^{\eta', x}(s)$ whenever $\mathscr{T}_t^{\eta'}(s, a)\leq 
\mathscr{T}_t^\eta(s, a)$.
for any $t$.  
Thus, by finding the risk-minimizing $\eta$, we can maximize the pool of safe policies.
Whenever we can, we shall therefore look not just for \textit{a} baseline policy that satisfies $\mathscr{T}^\eta(s_0) \leq  x_c^*$, but also for \textit{the} threat minimizing policy.
Lastly, if $\eta$ is $x^*_c$-safe, then $\eta \in \prod^{\eta, x^*_c}_1$ is guaranteed so that the  $\eta \in \prod^{\eta, x^*_c}_1$ is not empty. Unless otherwise denoted.  we will use $\prod^{\eta}$ to denote $\prod^{\eta, x^*_c}_1$, and use  $S^{\eta}, A^{\eta}(s)$ to denote $S^{\eta,\mathbf{x}^*_c}, A^{\eta, \mathbf{x}^*_c}(s)$.

\section{Reconnaissance-MDP (R-MDP) and Planning-MDP (P-MDP) } \label{sec:defRP}

As stated in the previous section, we can obtain a set of safe policy from any baseline policy $\eta$ satisfying $\mathscr{T}^\eta(s_0) \leq x_c^*$ , and that we can maximize this set by constructing the set of safe policy from the threat minimizing $\eta$. 
The purpose of R-MDP is thus to \textit{recon} the system prior to the reward maximization and look for the policy $\eta^*$ with minimal threat (maximal  $\prod^{\eta^*}$). %
As a process, R-MDP is same as the original MDP except that we have a danger function instead of a reward function, and the goal of the agent in the system is to find the minimizer of the risk: 
$\eta^*(s,a) :=  \argmin_\eta  \mathscr{T}^{\eta}(s,a)$.
This is, indeed, more than what we need when the safety is our only concern. 
So long that $\mathscr{T}^{\eta}(s,a) \leq  x^*_c$, the pool of policies $\pi \in \prod^{\eta}$ is guaranteed safe. 
The purpose of Planning-MDP (P-MDP) is to search within $\prod^{\eta}$ for a good reward-seeking policy.

P-MDP is the same as original MDP, except that action set $A$ is state and time dependent; that is, the agent is allowed to take action only from $A^{\eta}$ whenever $s \in S^\eta$, and take the deterministic action $\argmin_a E_\eta[\mathcal{T}_t (s,a)]$ whenever $s \not \in S^\eta$.

The purpose of P-MDP is to find the policy 
\begin{align}
\pi^* :=  \argmax_{\pi \in \prod^{\eta}} E_\pi \left[ \sum \gamma^t r_t \right].
\end{align}

The following algorithm will find a safe good policy if $\mathcal{T}^{\eta*} \leq  x_c^*. $ 

\begin{algorithm}[ht]  %
\caption{RP-algorithm}
\label{alg:RPalgorithm}
\begin{algorithmic}[1]

 \State Obtain $\eta^*(s,a) :=  \argmin_\eta  \mathscr{T}^{\eta}(s,a)$ defined in Eq.\eqref{eq:def risk function} for any $(s,a)$ or prepare a heuristically selected $\eta^*(s,a)$ .
 \State Evaluate $\mathscr{T}_t^{\eta^*}$ and Construct $\prod^{\eta^*}$ 
  \State Obtain $\pi^* :=  \argmax_{\pi \in  \prod^{\eta,x^*_c }} E_\pi \left[ \sum \gamma^t r_t \right]$ using either model-free or model-based RL.
\end{algorithmic}
\end{algorithm}

\subsection{Variants of Reconnaissance and Planning Algorithm}
In what follows, we describe important variants of the RP-algorithm that are useful in practice. 
\subsubsection{Constraint on the probability of fatal accident}
So far, we have considered the constraints of the form $E_{\pi}\left[\textstyle \sum_{k=1}^{T} \beta^k d_k \right] \leq c$.
If the danger to be avoided is so catastrophic that \textit{one accident alone} is enough to ruin the project, one might want to directly constrain the probability of an accident. 
Our RP-Algorithm can be used to find a safe solution for a CMDP with this type of constraint as well.
Let us use $d_t=d(s_{t-1},a_{t-1})$ to represent the binary indicator that takes the value 1 only if the agent encounters the accident upon taking the action $a_{t-1}$ at the state $s_{t-1}$.
Using this notation, we can write our constraint as $P( \max_{1\leq t \leq T} d_t = 1) \leq c$, and our threat function for this case can be recursively defined as follows:
\begin{align}
\mathscr{T}^{\pi}_{t}(s_{t}, a_{t}) 
=& E_{\pi}\left[ d_{t+1} 
+ (1-d_{t+1})E_{\pi}\left[\mathscr{T}^{\pi}_{t+1}(s_{t+1}, a_{t+1})\right]
\mid  s_{t}, a_{t} \right].
\end{align}
Notice that this is a variant of the Bellman relation for the original $\mathscr{T}^\eta$ \eqref{eq:def risk function} in which $\beta$ is replaced with  
$E_{\pi}\left[ 1-d_{t+1} \mid  s_{t}, a_{t}  \right]$.
With straight-forward computations, it can be verified that theorem \ref{thm:safe} follows if we replace  $\beta$ in the statement with the maximum possible value of $E_{\pi}\left[ 1-d_{t+1} \mid  s_{t}, a_{t}  \right]$. 
We can replace $\beta$ with its upperbound ($=1.0$) as well. 
That is, we may construct a set of safe policies by setting $x = \frac{c}{1 + (T-1)z}$.
With this strict constraint, however, $x$ approaches 0 as $T$ approaches infinity. 
This is in agreement with the law of large numbers; any accident with non-zero probability will happen almost surely if we wait for infinite length of time.

\subsubsection{Constraint on the probability of multiple fatal accidents}

Many application of CMDP involves multiple fatal events.  
For example, during the navigation of highway with heavy traffic, the driver must be wary of the movements of multiple other cars.
Industrial robots in hazardous environment might also have to avoid numerous obstacles. 

Our setup in the previous subsection can be used to find a solution for this type of problem.
Let us consider a model in which the full state of the system is given by $(s^{(o)}_t, \{s^{(n)}_{t}; 1,...,N\})$, where the state of the $n$-th obstacle is $s^{(n)}_{t}$ and the aggregate state of all other objects in the system is $s^{(o)}_{t}$(i.e. the location of the agent, etc). 
Let $\mathscr{T}^{\pi, n}_t(s^{(o)}_t,s^{(n)}_{t})$ 
be the probability of collision in the subsystem containing only the agent and the $n$-th obstacle.
Under this set up, we can appeal to a basic property of probability measure regarding a union of events to obtain the following interesting result:
\begin{theorem} \label{thm:sum_bound}
Let us assume that agent can take action based solely on $s^{(o)}_t$, and that $s^{(n)}_{t}$ are all conditionally independent of each other given $s^{(o)}_t$.  Then 
\begin{align} 
    \mathscr{T}_t^{\pi} (s_t,a_t) \leq  \sum_{n=1}^N \mathscr{T}^{\pi, n}_t(s^{(o)}_t,s^{(n)}_{t}).
\end{align} \label{eq:sum_bound}
\end{theorem} \vspace{-3mm}
We can then follow the procedure described in Section \ref{sec:method_core} with this new threat function to  construct the set of secure policies in  Corollary \ref{ref:thm2}. 

Theorem \ref{thm:sum_bound} is closely related to the risk potential approaches \cite{Wolf08RiskPotential, Rasekhipour16RiskPotential, Ji16RiskPotential}.%
These methods also work by evaluating the risk of collision with each obstacle for each location in the environment and by
superimposing the results.
However, most of them define the risk potential heuristically.

\section{Experiment}
We conducted a series of experiments to find answers to the following set of questions: 
\begin{enumerate}
    \item What does the threat function obtained from our method look like? 
    \item  How effective is the safe policy obtained from our method when suboptimal policy was used for the baseline policy $\eta$?
    \item  How well does the policy trained by our RP-method perform in new environments?
\end{enumerate}
We compared our algorithm's results on these experiments against those of other methods, including (1) classical MPC, (2) DQN with Lagrange penalty, and (3) Constrained Policy Optimization (CPO) \cite{Achiam17ConstrainedPO}.  %
At every step, the version of MPC we implemented in our study selects the best reward-seeking action among the set of actions that were deemed safe by the lookahead search. 
DQN with Lagrange penalty is a version of DQN for which the reward is penalized by the risk function with Lagrange weight.  
We tested this method with three choices of Lagrange weights. 
As for CPO, we used the implementation available on Github \cite{CPOcode}.

For our method, we used a neural network to approximate the threat function in the R-MDP for a heuristically-chosen baseline policy that is not necessarily safe, and solved the P-MDP with DQN.  %
When solving P-MDP with DQN, it becomes cumbersome to compute $Q^*$ on states outside $S^\eta$.   
We therefore constructed an MDP defined on $S^\eta$ that is equivalent to the original MDP for the policies in $\prod_1^{\eta,x_s}$.
Namely, we constructed the tuple $(S^{\eta}, A^{\eta}, r^{\eta}_P, P^{\eta}_P, P_0)$, whose components are defined as follows. 
The function $r^{\eta}_P( \cdot|s) $ is the restriction of the reward $r$ to $A^{\eta}(s)$ for all $s \in S^\eta$. %
$P^{\eta}_P$ is a transition probability function derived from the original state transition probability $P$ such that, for all $s_1, s_2 \in S^\eta$,  $ P^{\eta}_P(s_2 | s_1, a) = P(s_2 | s_1, a) + P ((s_1,a) \overset{(S^\eta )^c}{\longrightarrow} s_2)  $ 
where $(s_1,a) \overset{(S^\eta)^c}{\longrightarrow} s_2$ is the set of all trajectories from $s_1$ to $s_2$ that (1) take a detour to $(S^\eta )^c$ at least once after taking the action $a$ at $s_1$, (2) take the action $a^*(s') = \arg \min_{a\in A} \mathscr{T}^{\eta} (s', a)$ for all $s' \in (S^\eta )^c$, and (3) lead to $s_2$ without visiting any other states in $S^\eta$.

\subsection{Tasks}
\begin{figure}[ht]
\centering
\includegraphics[width=110mm]{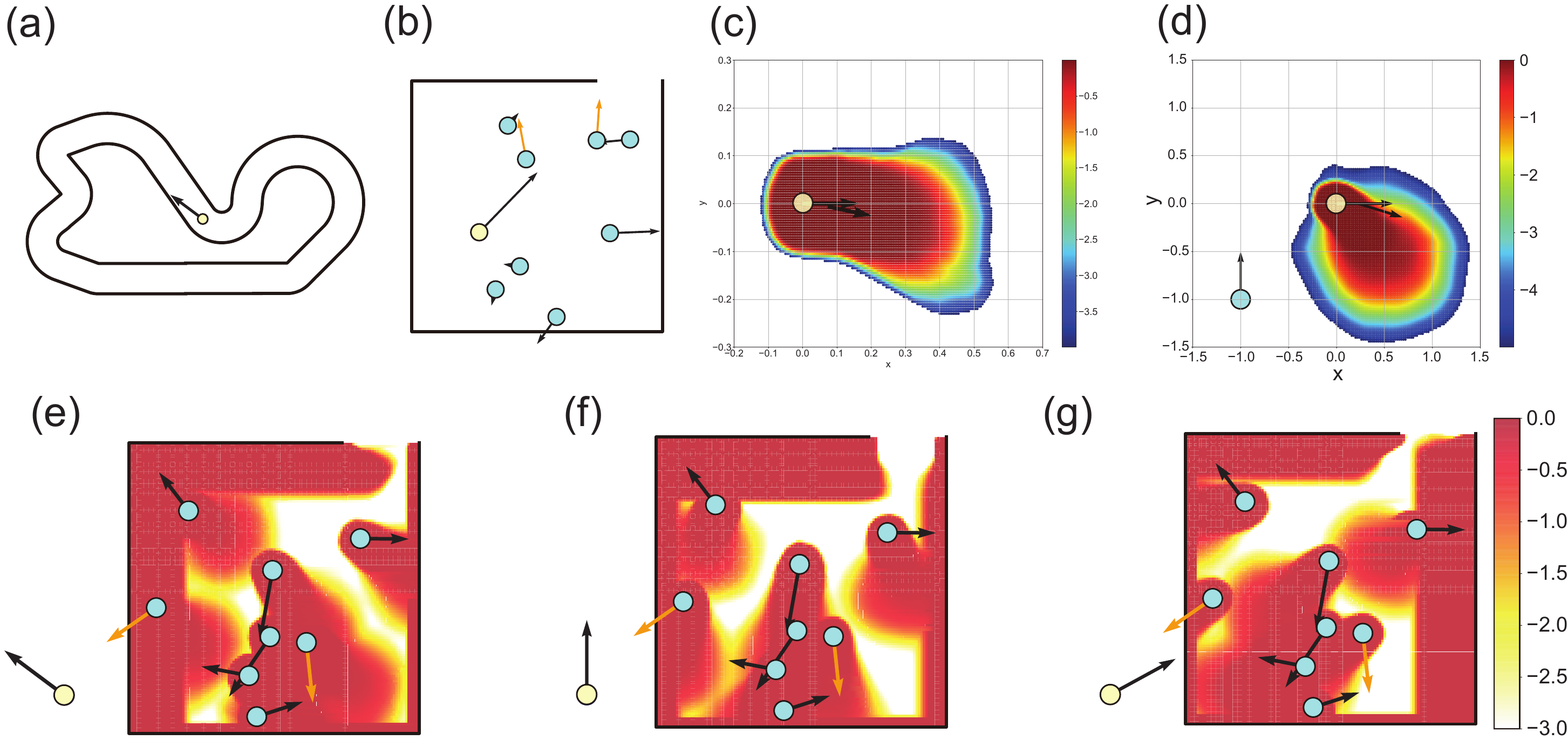}
\caption{Panels (a) and (b): the fields for Circuit task and Jam task. For Jam, the light blue circles are obstacles, and the yellow circle is the agent. The arrow attached to each object shows its direction of movement. 
Panels (c) and (d): the heat maps of the trained threat function in the neighborhood of the agent. The shape of the heat map changes with the speed and the direction of the object's movement.
(e),(f),(g) are the heat maps of the upper bound of the threat function (Theorem  \ref{thm:sum_bound}), computed for different velocity settings of the agent. The assumed movement of the agent is indicated at the left bottom corner of each map.  
These maps can be interpreted as a heat map of risk potential.}
\label{fig:field}
\end{figure}

We conducted experiments on three tasks on 2-D fields (see Fig. \ref{fig:field} for visualizations).
\begin{description}
    \item[Point Gather] In this task, the agent's goal is to collect as many green apples as possible while avoiding red bombs. 
    \item[Circuit] The agent's goal is to complete one lap around the circuit without crashing into a wall.  The agent can control its movement by regulating its acceleration and steering. Lidar sensors are used to compute the distance to obstacles.
    \item[Jam] The agent's goal is to navigate its way out of a room from the exit located at the top right corner as quickly as possible without bumping into 8 randomly moving objects.
\end{description}

\subsection{Learning the threat function}
The state spaces of Jam and Circuit are high-dimensional, because there are multiple obstacles in the environment that must be avoided.  
We therefore used the method described in Şection \ref{sec:defRP} to construct an upper bound for the true threat function by considering a set of separate R-MDPs in which there is only one obstacle. 
We also treated \textit{wall} as a set of immobile obstacles so that we can construct the threat function for the circuit of any shape. 
For more detail, see the appendix.%
Fig. \ref{fig:field} is the heat map for the upper bound of the threat function computed in the way of Theorem  \ref{thm:sum_bound}.  
Note that the threat map changes with the state of the agent.
We see that our threat function is playing a role similar to the risk potential function. 
Because our threat function is computed using all aspects of the agent's state (acceleration, velocity, location),  we can provide more comprehensive measure of risk in high dimensional environments compared to other risk metrics used in applications, such as TTC (Time To Collision) \cite{Lee76TTC} used in smart automobiles that considers only 1D movement.

\subsection{Learning performance}
\begin{figure}[ht]
\centering
\includegraphics[width=120mm]{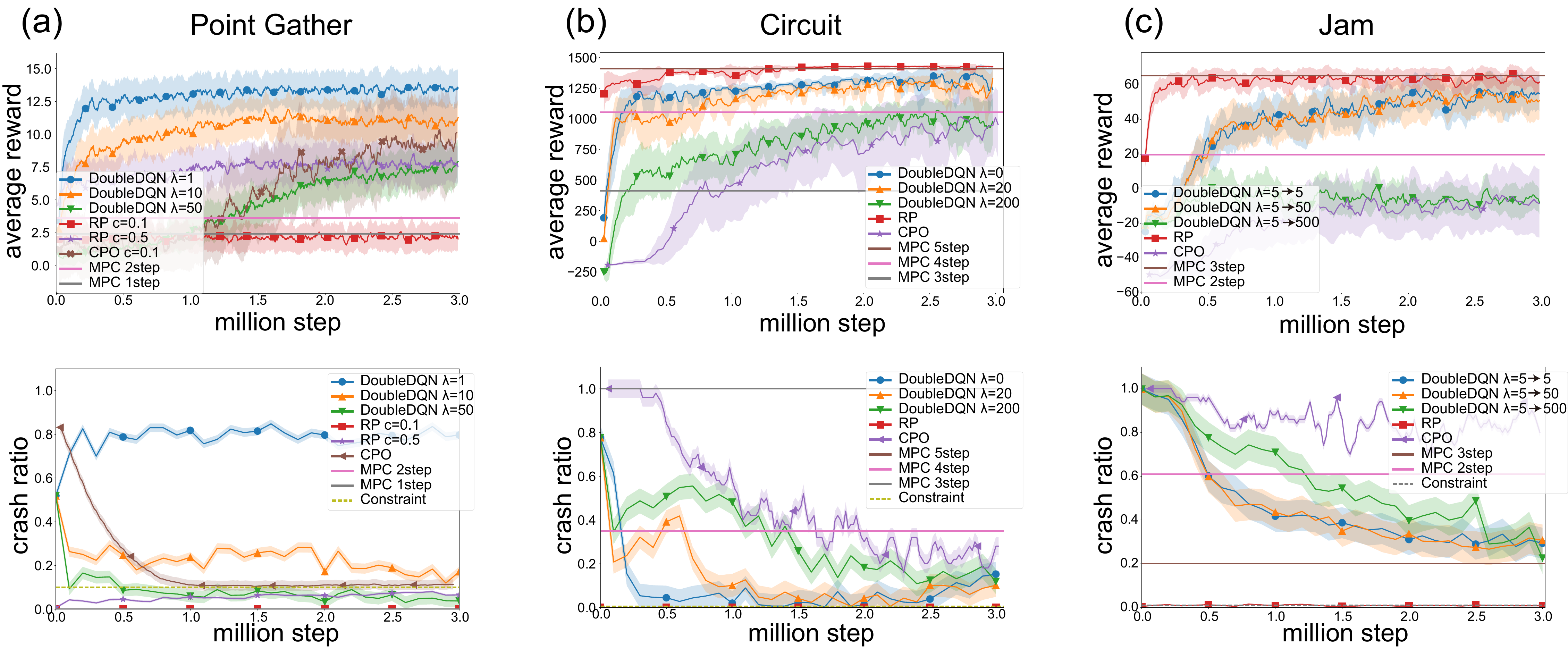}
\caption{Comparison of multiple CMDP methods in terms of rewards and crash rate.  For both Circuit and Jam, our method (P-DMP) achieves the highest average reward and lowest crash rate throughout the training process. 
DQN performs better in terms of reward for Point-Gather, but at the cost of a very high crash rate.
}
\label{fig:crash_rate}
\end{figure}

Fig. \ref{fig:crash_rate} plots the average reward and the crash rate of the policy against the training iteration for various methods.
The curve plotted for our method (RP) corresponds to the result obtained from training on the P-MDP.
The average and the standard deviation at each point was computed over 10 seeds.  %
As we can see in the figure, our method achieves the highest reward at almost all phases during the training for both Jam and Circuit, while maintaining the lowest crash rate. 
In particular, our method performs significantly better than other methods both in terms of safety and average reward for Jam, the most challenging environment. 
The RP-trained policy can safely navigate its way out of the dynamically-changing environment consistently even when the number of randomly moving obstacles is different than in the R-MDP used to construct the secure set of policies.

Penalized DQN performs better than our method in terms of reward for the Point-Gather, but at the cost of suffering a very high crash rate ($\sim 0.8$). 
Our method is also safer than the 3-step MPC for both Jam and Circuit as well, a method with significantly higher computational cost.  %

\subsection{Robustness of the learned policy to the change of environment}
We conducted two sets of experiments of applying a policy learned on one environment to the tasks on another environment. 
For the first set of experiments, we trained a safe policy for the circuit task, and evaluated its performance on the circuit environments that are different from the original circuit used in the training of the policy: 
(1) narrowed circuit with original shape, and (2) differently shaped circuit with same width. 
For the second set of experiments, we trained a safe policy for the JAM task, and tested its performance on other JAM tasks with different numbers of randomly moving obstacles.

Fig. \ref{Circuit2} and Fig. \ref{JAM2} shows the results.
For the modified Jam, we have no results for MPC with more than 3-step prediction since the search cannot be completed within reasonable time-frame. 
The 4-step MPC requires 36.5secs per episode (200 steps) for Circuit, and the 3-step MPC requires 285secs per episode (100 steps) for the original Jam.  %
We find that, even in varying environments, the policy obtained by our method can guarantee safety with high probability while seeking high reward. 

\begin{table}[h!]
    \centering
    \small
    \begin{tabular}{c||c|c|c|c}
         Environment & RP & MPC 4step & DQN $\lambda$=0 & DQN $\lambda$=200  \\ \hline
         Training env. & 1439 (0) & 1055 (0.35) & 1432 (0.05) & 933 (0.4) \\ \hline
         Narrowed env. & 377 (0) & 959 (0.55) & \textcolor{red}{-151} (1.0) & \textcolor{red}{-145} (0.99) \\ \hline
         Circle & 130 (0) & 351 (0) & \textcolor{red}{-189} (1.0) & \textcolor{red}{-171} (1.0) \\ \hline
         Computation Time (s) & 1.0 & 36.5 & 0.9 & 0.9   \\ \hline 
        \end{tabular}
    \caption{Performance of trained policies on unknown Circuit environments. The values in the table are the obtained rewards, with the probabilities of crashing within parentheses. The agent was penalized 200 pts for each collision, and rewarded for the geodesic distance traveling along the course in right direction. For details concerning the reward settings, please see the appendix.}%
    \label{Circuit2}
\end{table}

\begin{table}[ht]
    \centering
    \small
    \begin{tabular}{c||c|c|c|c|c}
         Environment & RP & MPC 2step & MPC 3step & DQN $\lambda$=5 & DQN $\lambda$=500 \\ \hline 
         3 obstacles & 78.2 (0) & 47.45 (0.33) & 77.5 (0.05) & 77.2 (0.04) & 4.4 (0.17) \\ \hline 
         8 obstacles (training env.) & 69.1 (0) & 21.32 (0.59) & 65.3 (0.2) & 47.1 (0.38) & -1.0 (0.24) \\ \hline
         15 obstacles & 33.0 (0.02) & -2.5 (0.8) & 36.6 (0.45) & 16.5 (0.66) & -16.8 (0.51) \\ \hline
         Computation Time (s) & 1.2 & 2.8 & 285 & 0.4 & 0.4   \\ \hline 
        \end{tabular}
    \caption{Performance of trained policies on and unknown Jam environments}
    \label{JAM2}
\end{table}

\section{Conclusion}

Our study is the first of its kind in providing a framework for solving CMDP problems that performs well in practice on high-dimensional dynamic environments like Jam.  %
Although our method does not guarantee finding the optimal reward-seeking safe policy, empirically, it is able to find a policy that performs significantly better than classical methods both in terms of safety and rewards. 
Our treatment of the threat function 
helps us obtain a more sophisticated and comprehensive measure of danger at each state than conventional methods.%
Our bound on the threat function also seem to have close connections with previous Lyapunov-based methods as well.
Overall, we find that utilizing threat functions is a promising approach to safe RL and further research on framework may lead to new CMDP methods applicable to complex, real-world environments.

\subsubsection*{Acknowledgments}
We thank Wesley Chung for his valuable comments on the manuscript and proposal of a tighter bound.

\newpage
\section{Appendix}
\subsection{Proof of Theorem \ref{lemma:threat bound}}
%
\begin{proof}
This bound can be proved by the recursion derived from the Bellman equation.
Define $e_t(s_t,a_t) \equiv \mathscr{T}^{\pi}_{t}(s_{t}, a_{t})-\mathscr{T}^{\eta}_{t}(s_{t}, a_{t})$.
Then $e_t(s_t,a_t)$ has the following recurrence.
\begin{align}
& e_{t-1}(s_{t-1},a_{t-1}) \nonumber \\
=&  \mathscr{T}^{\pi}_{t-1}(s_{t-1}, a_{t-1})
- \mathscr{T}^{\eta}_{t-1}(s_{t-1}, a_{t-1})\nonumber \\
=& \beta E\left[E_{\pi}\left[ \mathscr{T}^{\pi}_t(s_t, a_t)\right]
-E_{\eta}\left[\mathscr{T}^{\eta}_t(s_t, a_t)\right]  \mid  s_{t-1}, a_{t-1} \right]
\nonumber \\
=& 
\beta E\left[E_{\pi}\left[ e_t(s_t, a_t)\right] 
+ E_{\pi}\left[\mathscr{T}^{\eta}_t(s_t, a_t)\right]
- E_{\eta}\left[\mathscr{T}^{\eta}_t(s_t, a_t)\right]
\mid  s_{t-1}, a_{t-1} \right].
\label{eq:recursion of e}
\end{align}
From this recursion \eqref{eq:recursion of e}, 
we can derive an inequality.
\begin{align}
& e_{t-1}(s_{t-1},a_{t-1}) \nonumber \\
=& 
\beta E\left[E_{\pi}\left[ e_t(s_t, a_t)\right] 
+ E_{\pi}\left[\mathscr{T}^{\eta}_t(s_t, a_t)\right]
- E_{\eta}\left[\mathscr{T}^{\eta}_t(s_t, a_t)\right]
\mid  s_{t-1}, a_{t-1} \right]
\nonumber \\
= & \beta  E\left[E_{\pi}\left[ e_t(s_t, a_t)\right]\mid  s_{t-1}, a_{t-1} \right]
\nonumber \\
& + \beta \int_{s_t \in S^{\eta}} P(s_t  \mid  s_{t-1}, a_{t-1}) (E_{\pi}\left[\mathscr{T}^{\eta}_t(s_t, a_t) \mid s_t \right]
- E_{\eta}\left[\mathscr{T}^{\eta}_t(s_t, a_t) \mid s_t \right] ) ds_t
\nonumber \\
& +\beta \int_{(s_t \in (S^{\eta})^c} P(s_t \mid  s_{t-1}, a_{t-1})
(E_{\pi}\left[\mathscr{T}^{\eta}_t(s_t, a_t) \mid s_t  \right]
- E_{\eta}\left[\mathscr{T}^{\eta}_t(s_t, a_t) \mid s_t \right])ds_t
\nonumber \\
\leq & \beta  E_{\pi}\left[ e_t(s_t, a_t)\mid  s_{t-1}, a_{t-1} \right]
 + \beta x_t E\left[1_{s_t \in S^{\eta}} y_t(s_t) \mid  s_{t-1}, a_{t-1} \right]
\end{align}
where
$y_t(s_t) = \sum_{a_t: \pi(a_t|s_t) > \eta(a_t|s_t)}
\pi(a_t|s_t)  - \eta(a_t|s_t) = d_{TV}(\pi(\cdot|s_t), \eta(\cdot|s_t))$.
Here we replaced $\mathscr{T}^{\eta}_t(s_t, a_t)$ in the 2nd term by its maximum value on the support of $\pi$ when $s_t \in S^{\eta}$ and replaced $\mathscr{T}^{\eta}_t(s_t, a_t)$ in the 3rd term by its minimum value, zero. 

For convenience, let $z_t(s_{t-1},a_{t-1}) = E_\theta\left[1_{s_t\in S^{\eta}}y_t(s_t) \mid  s_{t-1}, a_{t-1} \right]$. Then the above inequality can be written as $e_{t-1}(s_{t-1},a_{t-1}) \leq  \beta  E_{\pi}\left[ e_t(s_t, a_t)\mid  s_{t-1}, a_{t-1} \right]
 + \beta x_t z_t(s_{t-1},a_{t-1})$.

Since $e_{T-1}(s_{T-1},a_{T-1}) = 0$ for the finite horizon MDP with length $T$, we have the inequality below by combining the above two inequalities and repeating the recursion.
\begin{align}
e_{0}(s_{0}, a_{0}) 
\leq & \beta  E_{\pi}\left[ e_1(s_1, a_1)\mid  s_{0}, a_{0} \right]
 + \beta x_1 z_1(s_{0},a_{0})
\nonumber \\
\leq & \beta^2  E_{\pi}\left[ e_2(s_2, a_2)\mid  s_{0}, a_{0} \right] + \beta^2 x_2 E_{\pi}\left[ z_2(s_1,a_1)\mid  s_{0}, a_{0} \right]
 + \beta x_1 z_1(s_{0},a_{0})
\nonumber \\
\leq & \sum_{t=1}^{T-1}\beta^t x_t E_{\pi}\left[ z_t(s_{t-1},a_{t-1})\mid  s_{0}, a_{0} \right]
\end{align}
This means
\begin{align}
\mathscr{T}^{\pi}_{0}(s_{0})
=&E_{\pi}\left[\mathscr{T}^{\pi}_{0}(s_{0}, a_{0}) \mid s_0 \right] \nonumber \\
=& E_{\pi}\left[\mathscr{T}^{\eta}_{0}(s_{0}, a_{0}) + e_{0}(s_{0}, a_{0}) \mid s_0 \right] \nonumber \\
\leq & x_0 + \sum_{t=1}^{T-1} \beta^t x_t E_{\pi}\left[ z_t(s_{t-1},a_{t-1})\mid s_0 \right].
\end{align}
Note this bound is tight in the sense that there exists some policy and environment that achieves this bound with equality.
To keep $z_T(s_{t-1},a_{t-1})$ small, we should make $y_t(s_t)$ small by making $\pi$ enough close to $\eta$.
For example, $\epsilon$-$\eta$ policy, i.e., take actions according to the policy $\eta$ with probability $1-\epsilon$ and take actions according to the uniform random policy $u$ with probability $\epsilon$, which becomes small when $\epsilon$ is small.
$y_t(s_t) = \frac{1}{2}\sum_{a_t \in A^{\eta}(s_t,t)}
|\pi(a_t|s_t)  - \eta(a_t|s_t)| 
\leq \sum_{a_t \in A^{\eta}(s_t,t)}
|(1-\epsilon)\eta(a_t|s_t)+\epsilon u  - \eta(a_t|s_t)| 
= \frac{1}{2}\epsilon\sum_{a_t \in A^{\eta}(s_t,t)} |\eta(a_t|s_t) - u|$.

\end{proof}

\subsection{Proof of Theorem \ref{thm:sum_bound}}
\setcounter{theorem}{1}
\begin{proof}
Let us suppose that we can write 
$s_t=(s^{(1)}_t,\cdots,s^{(N)}_t,s^{(o)}_t)$,
and let us denote the accident of $k$th type by $S_k $.    
Then, by a basic property of probability distribution 
\begin{eqnarray}
 P(\cup _{i=1}^N S_i | \cdot) \leq \sum_{i=1}^N P(S_i | \cdot).
\end{eqnarray}
Thus, if the transition probability is given by
$p(s_{t+1}|s_t,a_t)=p(s^{(o)}_{t+1}|s^{(o)}_t,a_t)
\prod_{n=1}^n p(s^{(n)}_{t+1}|s^{(n)}_{t},s^{(o)}_t,a_t)$ and if the policy being followed is $\pi(a_t | s^{(o)}_t)$, 
We have
\begin{eqnarray}
 P(\cup _{i=1}^N S_i | s_t, a_t) \leq \sum_{i=1}^N P(S_i | s^{(o)}_t,s^{(n)}_{t}, a_t).
\end{eqnarray}
Now, let us consider the problem of constraining the probability of an accident 
$P( \max_{1\leq t \leq T} d_t = 1 | \pi)$. 
Then,  using the fact that 
$\mathscr{T}_t^{\pi} (s_t, a_t) = P( \max_{1\leq t \leq T} d_t = 1 | s_t, a_t, \pi)$, we see that we can bound the threat by 
\begin{align} 
    \mathscr{T}_t^{\pi} (s_t,a_t) \leq  \sum_{n=1}^N \mathscr{T}^{\pi, n}_t(s^{(o)}_t,s^{(n)}_{t}), 
\end{align}
where $d^{(n)}_t (s^{(o)}_t,s^{(n)}_{t},a_t)$ is the 
probability of the collision with $n$-th obstacle at state $(s^{(o)}_{t+1},s^{(n)}_{t})$ in the 'sub'system consisting of only agent and $n$-th obstacle, and $\mathscr{T}^{\pi, n}_t(s^{(o)}_t,s^{(n)}_{t})$ is the threat function for the R-MDP on such subsystem.

\end{proof}

\subsection{Environments}
\subsubsection*{Circuit}
In this task, the agent's goal is to complete one lap around the circuit without crashing into the wall.
Each state in the system was set to be the tuple of (1) location, (2) velocity, and (3) the direction of the movement.
The set of actions allowed for the agent was 
$\{$0.15rad to left, 0.05rad to left, stay course,  0.05 rad to right, 0.15 rad to right $\} \times \{$ 0.02 unit acceleration, no acceleration,   0.02 unit deceleration $\}$ (15 options).   
At all time, the speed of the agent was truncated at $\{-0.04, 0.1 \}$. 
We rewarded the agent for the geodesic distance traveling along the course during each time interval, which accumulates to $1250$pts for one lap 
while we gave negative rewards for the stopping and collision during the time step, each of which are $-1$pts and $-200$pts, respectively.
We set the length of the episode to 200 steps, which is the approximate number of steps required to make one lap. 

\subsubsection*{Jam}
In this task, the agent's goal is to navigate its way out of a room from the exit located at the top left corner without bumping into 8 randomly moving objects. 
We set three circular safety zones centered at each corner except for the top left corner, i.e., exit.
Any moving obstacle entered into the safety zone disappear.
Without the safety zone, the task seems to be too difficult, i.e., there is a situation that the agent cannot avoid the collision even if the agent tried his best.
We set the safety zone to ease the problem hoping the probability that the agent can solve the task when employing the optimal policy becomes reasonably high.
The field was $3 \times 3$ square and the radius of the safety zone located at three corners were set to 0.5.
The radius of the agent and moving obstacles were set to 0.1.
We rewarded the agent for its distance from the exit, and controlled its value so that the accumulated reward at the goal will be around $85$. The agent was given $10$ points when it reaches the goal, was penalized $-0.05$ points for stopping the advance, and was given $50$pts penalty for each collision.
Similar to the setting as in Circuit, the agent was allowed to change direction and acceleration of the movement simultaneously at each time point.
The set of actions allowed for the agent was
$\{$0.30 rad to left, 0.10 rad to left, stay course,  0.10 rad to right, 0.30 rad to right $\} \times \{$ 0.02 unit acceleration, no acceleration,  0.02 unit deceleration $\}$.
At all time, the speed of the agent was truncated at $\{-0.1, 0.1 \}$. 
Each obstacle in the environment was allowed to take a random action from  $\{$0.15rad to left, 0.05rad to left, stay course,  0.05 rad to right, 0.15 rad to right $\} \times \{$ 0.02 unit acceleration, no acceleration,   0.02 unit deceleration $\}$.
The speed of the environment was truncated at $\{0, 0.06\}$.
We set the length of each episode to $100$ steps. 

\subsubsection*{Point Gather}
In this task, the goal is to collect as many green apples as possible while avoiding red bombs. 
There are 2 apples and 10 bombs in the field.
The agent was rewarded 10pts when the agent collected apple, and also reward
The point mass agent receives 29-dimensional state and can take two-dimensional continuous actions.
The state variable takes real values including, position and velocity, the direction and distance to the bomb, etc.
The action variables determine the direction and the velocity of the agent.
For the implementation of DQN, we discretized each action variable into 9 values.
We used the exact same task setting as the one used in the original paper.

\subsection{Model architecture and optimization details}
\subsubsection{RP-algorithm}
We implemented our method on Chainer \cite{Tokui15Chainer}.
\subsubsection*{Learning of Threat function} 
For the Circuit and Jam tasks, the agent must avoid collisions with both moving obstacles and the wall.
For these environments, it is computationally difficult to obtain the best $\eta$. 
Thus, we computed a threat function for the collision with every object individually in the environment under an arbitrary baseline policy $\eta$, and constructed the pool of approximately safe policies using the the upper bound of the threat function computed in the way we described in Section 3.1.2. 
We used $\eta$ that (1) decides the direction of the movement by staying course with probability $0.6$, turning right with probability $0.2$ and turning left with probability $0.2$, and (2) decides its speed by accelerating with probability $0.2$ and decelerating with probability $0.2$.  
Then, we trained two threat functions 
each of which predict the collision with the immobile point and moving obstacle randomly put in the 2D region.
Since $\eta$ is fixed, the threat function can be obtained by supervised learning, in which the task is to predict the future collision when starting from a given current state and employing the policy $\eta$.
The threat function for the collision with the immobile point is used to avoid collision with the wall, which can be considered as a set of immobile points.
We shall emphasize that, with our method described in Section 3, the environment used in P-MDP is different from the environment used in R-MDPs, because each threat function in the summand of \eqref{eq:sum_bound} is computed on the assumption that there is only one obstacle, either immobile point or moving obstacle in the environment.

We used neural network with three fully connected layers (100-50-15) and four fully connected layers (500-500-500-15) for the threat function of the immobile point and moving obstacle, respectively.
For the training dataset, we sampled 100,000 initial state and simulated 10,000 paths of length $5$  from each initial state.
We trained the network with Adam. 
The parameter settings for the training of the threat function of immobile point and of mobile obstacle are ($\alpha =1e-2, \epsilon =1e-2$,  batchsize = 512, number of epochs = 20), and ($\alpha =1e-2, \epsilon =1e-2$,  batchsize = 512, number of epochs = 25), respectively. 

For the point gather task, we again used the upper bound approximation explained in Section 3.1.2 for the threat function. The threat function is estimated by using a two-layer fully connected neural network.

\subsubsection*{Solving P-MDP} 
For the Planning MDP, we used DQN.
For DQN, we used convolutional neural network with one convolutional layer and three fully connected layers, 
and we trained the network with Adam ($\alpha =1e-3, \epsilon =1e-2$). 
We linearly decayed the learning rate from $1$ to $0.05$ over 3M iterations.

\subsubsection{DQN with Lagrange coefficient}
For DQN, we used the identical DQN as the one used for P-MDP.
For Lagrange coefficient, we tested with three Lagrange coefficients for each task, $\{0, 20, 200 \}$ for Circuit, $\{5, 50, 500 \}$ for Jam, $\{1, 10, 50 \}$ for Point Gather, respectively.
For the Jam task, the initial Lagrange coefficients are all set to 5 and gradually incremented to the final values $\{1, 10, 50 \}$ as done in \cite{Miyashita18MobileRobot}.
This heuristic pushes the agent to learn the goal-oriented policy first, and then learn the collision avoidance.

\subsubsection{Constrained Policy Optimization}
As for CPO, we used the same implementation publicized on Github \cite{CPOcode}, i.e., the policy is a Gaussian policy implemented by a two layer MLP with hidden units 64 and 32 for all tasks.

\subsubsection{Model Predictive Control}
We tested MPC with receding horizon $\in \{3,4,5\} $, and at every step selected the action with the highest reward among those that were deemed safe by the prediction.

\subsection{Related works}
CPO \cite{Achiam17ConstrainedPO} is a method that gradually improves the safe policy by making a local search for a better safe policy in the neighborhood of the current safe policy.
By nature, at each update, CPO has to determine \textit{what members of the neighborhood of current safe policy} satisfy the safety constraint. 
In implementation, this is done by the evaluation of Lagrange coefficients. 
Accurate selection of \textit{safe} policy in the neighborhood is especially difficult when the danger to be avoided is "rare" and "catastrophic", we would need massive number of samples to verify whether a given policy is safe or not.
Moreover, because each update is incremental, they have to repeat this process multiple times (usually, several dozens of times for Point Circle, and several thousands of times for Ant Gather and Humanoid Circle).  
Lyapunov based approach \cite{Chow18Lyapunov, Chow19Lyapunov} 
are also similar in nature.
At each step of the algorithm, Lyapunov based approach construct a set of safe policy from a neighborhood of the baseline policy, and computes from its safety-margin function---or the state-dependent measure of how bold an action it can take while remaining safe---to specify the neighborhood from which to look for better policy. 
For the accurate computation of the margin,  one must use the transition probability and solve the linear programming problem over the space with dimension that equals the number of states. 
The subset of the neighborhood computed from approximate margin may contain unsafe policy. 
Model checking is another approach to guarantee the safety.
Once the constraints are represented in the form of 
temporal logic constrains or computation-tree logic \cite{Baier03ModelCheck,Wen15TLC,Bouton19TLC},
we could ensure the safety by using model checking systems.
However, it is sometimes difficult to express the constraints such a structured form. Also even when we represents the constraints in the structured form, we again encounters computation issues; when the state-action space becomes large, the computation required for the model checking system becomes prohibitively heavy due to the increase of the candidates of the solutions.

\bibliography{main_nips}
\bibliographystyle{plain}

\end{document}